\documentclass[letterpaper, 10 pt, conference]{ieeeconf}  

\IEEEoverridecommandlockouts                              
\usepackage[T1]{fontenc} 
\usepackage{booktabs} 
\usepackage{hyperref}
\usepackage{graphicx} 
\usepackage{subfigure}
\usepackage{multirow} 
\usepackage{makecell} 
\usepackage{amsfonts} 
\usepackage[sorting=none, style=ieee]{biblatex}
\usepackage[linesnumbered,ruled,commentsnumbered]{algorithm2e}
\usepackage{amsmath} 
\title{\LARGE \bf
MGTraj: Multi-Granularity Goal-Guided Human Trajectory Prediction with Recursive Refinement Network
}

\author{Ge Sun and Jun Ma, \textit{Senior Member, IEEE}
\thanks{Ge Sun and Jun Ma are with the Division of Emerging Interdisciplinary Areas, The Hong Kong University of Science and Technology, Hong Kong SAR, China, and the Robotics and Autonomous Systems Thrust, The Hong Kong University of Science and Technology (Guangzhou), Guangzhou, China  (e-mail: gsunah@connect.ust.hk; jun.ma@ust.hk).}
}

\begin{document}

\newcommand{\etal}{\textit{et al.}} 

\maketitle
\thispagestyle{empty}
\pagestyle{empty}

\begin{abstract}
Accurate human trajectory prediction is crucial for robotics navigation and autonomous driving. Recent research has demonstrated that incorporating goal guidance significantly enhances prediction accuracy by reducing uncertainty and leveraging prior knowledge. Most goal-guided approaches decouple the prediction task into two stages: goal prediction and subsequent trajectory completion based on the predicted goal, which operate at extreme granularities: coarse-grained goal prediction forecasts the overall intention, while fine-grained trajectory completion needs to generate the positions for all future timesteps. The potential utility of intermediate temporal granularity remains largely unexplored, which motivates multi-granularity trajectory modeling. While prior work has shown that multi-granularity representations capture diverse scales of human dynamics and motion patterns, effectively integrating this concept into goal-guided frameworks remains challenging.
In this paper, we propose MGTraj, a novel \textbf{M}ulti-\textbf{G}ranularity goal-guided model for human \textbf{Traj}ectory prediction. MGTraj recursively encodes trajectory proposals from coarse to fine granularity levels. At each level, a transformer-based recursive refinement network (RRN) captures features and predicts progressive refinements. Features across different granularities are integrated using a weight-sharing strategy, and velocity prediction is employed as an auxiliary task to further enhance performance. Comprehensive experimental results in EHT/UCY and Stanford Drone Dataset indicate that MGTraj outperforms baseline methods and achieves state-of-the-art performance among goal-guided methods.

\end{abstract}
\section{Introduction}
Human trajectory prediction is an essential yet challenging task in various applications, including autonomous driving, social robot navigation and planning, and surveillance systems. By inferring human future motion based on historical observation, the autonomous vehicle or the social-compliant robots can generate reasonable path to proactively avoid collisions with surrounding humans, hence enhancing the safety of the system.
Early deep learning-based methods primarily focus on forecasting human trajectories in a straightforward manner~\cite{sociallstm_2016_CVPR, SocialGAN_2018_CVPR, socialattention_2018_ICRA, Trajectron++_2020_ECCV}, despite the high degree of uncertainty in this task hinders precise prediction. Considering that human motion is inherently goal-driven, researchers introduce goal-guided methods to reduce the uncertainty of the trajectory prediction task, and it has been proven that accurately predicting the goal is beneficial in improving the accuracy of trajectory prediction~\cite{DenseTNT_2021_ICCV, Goal-GAN_2020_ACCV}.

\begin{figure}[t]
    \scriptsize
    \setlength{\tabcolsep}{1.5pt}
    \centering
    \includegraphics[width=1\linewidth]{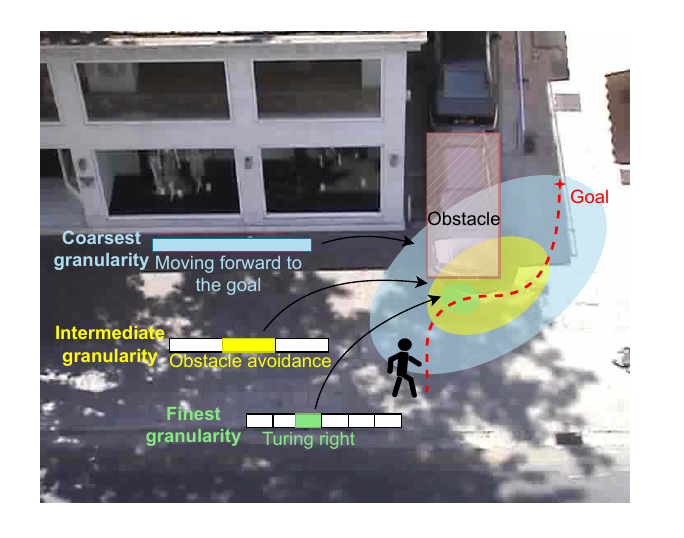}
    \\
    \caption{Illustration of human motion intention when observed from different temporal granularities. The colored block indicates the observation window, and different colors represent different granularities. The intermediate temporal granularity (in yellow) could be beneficial to understand human trajectories.}
    \label{fig_1}
    \vspace{-14pt}
\end{figure}

Previous goal-guided human trajectory prediction methods typically adopt a two-stage framework: goal prediction and trajectory completion. In the goal prediction stage, various networks are applied to predict the endpoint, including CNN~\cite{Goal-GAN_2020_ACCV, YNet_2021_ICCV, GoalSAR_2022_CVPRW, GDTS_IROS_2025}, RNN~\cite{GTP_WACV_2021, SGNet_RAL_2022}, MLP~\cite{PECNet_2020_ECCV, Bitrap_RAL_2021}, GCN~\cite{ControlPoint_AAAI_2023}, and Transformer~\cite{PPT_ECCV_2024}, either based on historical observation alone or additional map information. Besides, Zhao~\etal~\cite{Expert_CVPR_2021} builds an expert repository to retrieve possible goals. Receiving the goal prediction results, in the trajectory completion stage, the entire future trajectory is predicted in a recurrent or simultaneous way. From the perspective of temporal granularity, the initial goal prediction stage identifies the coarsest representation of global human intention, typically expressed as a target position or probability distribution. Conversely, the trajectory completion task aims to forecast the future trajectory at the finest resolution, requiring the prediction of positions for all future timesteps within the prediction horizon.

While goal-guided prediction methods have explored leveraging coarse-grained goal information to enhance fine-grained trajectory forecasting, the potential utility of intermediate temporal granularity remains largely unexplored. This intermediate level may capture richer behavioral semantics, such as tactical maneuvers or navigational strategies. As illustrated in Fig. \ref{fig_1}, human motion exhibits distinct characteristics across granularity levels: at the coarsest level in blue, movement is directed towards a destination; even with a deterministic goal, the path can exhibit stochasticity. Hence, instantaneous kinematics (e.g., turning right) are captured at the finest level. However, the underlying causal factors motivating such kinematic maneuvers (e.g., why a right turn occurs) remain ambiguous without observing dynamics at an intermediate granularity. Critically, this intermediate level reveals higher-level semantics, such as obstacle avoidance tactics. Explicitly modeling these intermediate-granularity dynamics offers significant potential for enhancing behavioral understanding and, consequently, improving prediction accuracy in trajectory forecasting.

Preliminary research has explored the application of intermediate granularity for trajectory prediction. TPNMS~\cite{TPNMS_AAAI_2021} is the first to model motion in multiple temporal granularities as temporal pyramids. The features of different granularities are averaged for fusion and then decoded to future trajectory with multiple granularities using the proposed multi-supervision training scheme. Alternatively, V$^2$-Net~\cite{V2_ECCV_2022} converts trajectories to frequency domains via Fourier transform, predicts spectrum keypoints, and reconstructs trajectories through interpolation, though its frequency-space representation lacks intuitive behavioral interpretation. MlgtNet~\cite{MlgtNet_TITS_2024} also adopts Gabor transforms for multi-granular feature extraction but ultimately regresses only finest-granularity trajectories, discarding explicit intermediate representations. Crucially, none of them coherently integrate intermediate granularity within goal-guided frameworks or leverage it to bridge coarse goals with fine-grained timesteps.

Motivated by this, we propose \textbf{MGTraj}, a goal-guided \textbf{M}ulti-\textbf{G}ranularity Human \textbf{Traj}ectory Prediction model. First, goal prediction is obtained leveraging a pretrained goal predictor, and then initial trajectory proposals are generated conditioned on the goal prediction. 
We then convert the entire trajectory (observed history coupled with initial trajectory proposal) into a pre-defined coarse granularity by granularity conversion. The converted trajectory is then fed into a transformer-based Recursive Refinement Network (RRN) to predict the refinement and update the future trajectory proposal. The updated trajectory proposal is subsequently converted into a finer granularity. This granularity conversion and refinement process is repeated until the finest-granularity prediction is obtained. To facilitate the fusion of features across different levels of granularity, we employ a weight-shared transformer encoder within the RRN, effectively training the model to encode trajectories of varying granularities while implicitly recognizing that these inputs represent the same trajectory. Furthermore, to enhance predictive performance, we introduce velocity prediction as an auxiliary task to incorporate soft constraints through joint optimization. The velocity is predicted and refined along with the position. By effectively capturing and integrating the features from different temporal granularities of trajectory, MGTraj contributes to the advancement of more accurate goal-guided human trajectory prediction, even without using map information. In summary, the key contributions of this work are listed as follows:
\begin{itemize}
\item First, we provide a thorough analysis on previous goal-guided pedestrian trajectory prediction work from the perspective of granularity. By formalizing goal prediction as the coarsest granular representation, we propose MGTraj, a novel goal-guided multi-granularity human trajectory prediction framework that explicitly models and captures features across multiple temporal granularity levels.
\item We propose a transformer-based recursive refinement network to hierarchically extract features across multi-granular temporal dynamics. To fuse these features, a weight-shared scheme is applied. Besides, we also introduce velocity augmentation and auxiliary velocity prediction task, enforcing soft kinematic constraints through joint optimization to improve trajectory prediction accuracy.
\item Extensive experiments are conducted across multiple public datasets, including ETH/UCY dataset and Stanford Drone Dataset (SDD), and the results demonstrate the state-of-the-art performance of our model.
\end{itemize}

\section{RELATED WORKS}
\label{sec:related works}

\subsection{Goal-Guided Human Trajectory Prediction}
To reduce uncertainty and increase the accuracy of human trajectory prediction, researchers have developed goal-guided methods that explicitly incorporate goal information into trajectory prediction. These approaches often decouple the entire trajectory prediction task into two subtasks: goal prediction and trajectory completion. For example, PECNet~\cite{PECNet_2020_ECCV} employs a variational autoencoder (VAE) to estimate multiple goal candidates and leverages social pooling for extracting relevant neighbor information.  Similar to it, BiTraP~\cite{Bitrap_RAL_2021} also adopts a VAE architecture for goal prediction, and a bi-directional GRU decoder is applied to generate trajectories conditioned on these predicted goals. Goal-SAR~\cite{GoalSAR_2022_CVPRW} adopts a recurrent architecture that predicts the position of each future frame using both goal predictions and all historical states leading up to that moment. Expert-GMM~\cite{Expert_CVPR_2021} builds a goal expert repository based on the training dataset, and then retrieves the goal estimation from this repository. More recently, GDTS~\cite{GDTS_IROS_2025} further integrates goal guidance with diffusion denoising models, leveraging a transformer-based denoising network and tree-sampling algorithm for trajectory completion. To capture both short-term dynamics and long-term dependencies, PPT~\cite{PPT_ECCV_2024} employs progressive pretraining:  first predicting the position of the next frame, and then predicting the final goal. The trajectory completion module subsequently leverages representations from both pretrained stages for final trajectory prediction.

\begin{figure*}[ht]
    \scriptsize
    \setlength{\tabcolsep}{1.5pt}
    \centering
    \includegraphics[width=0.95\linewidth]{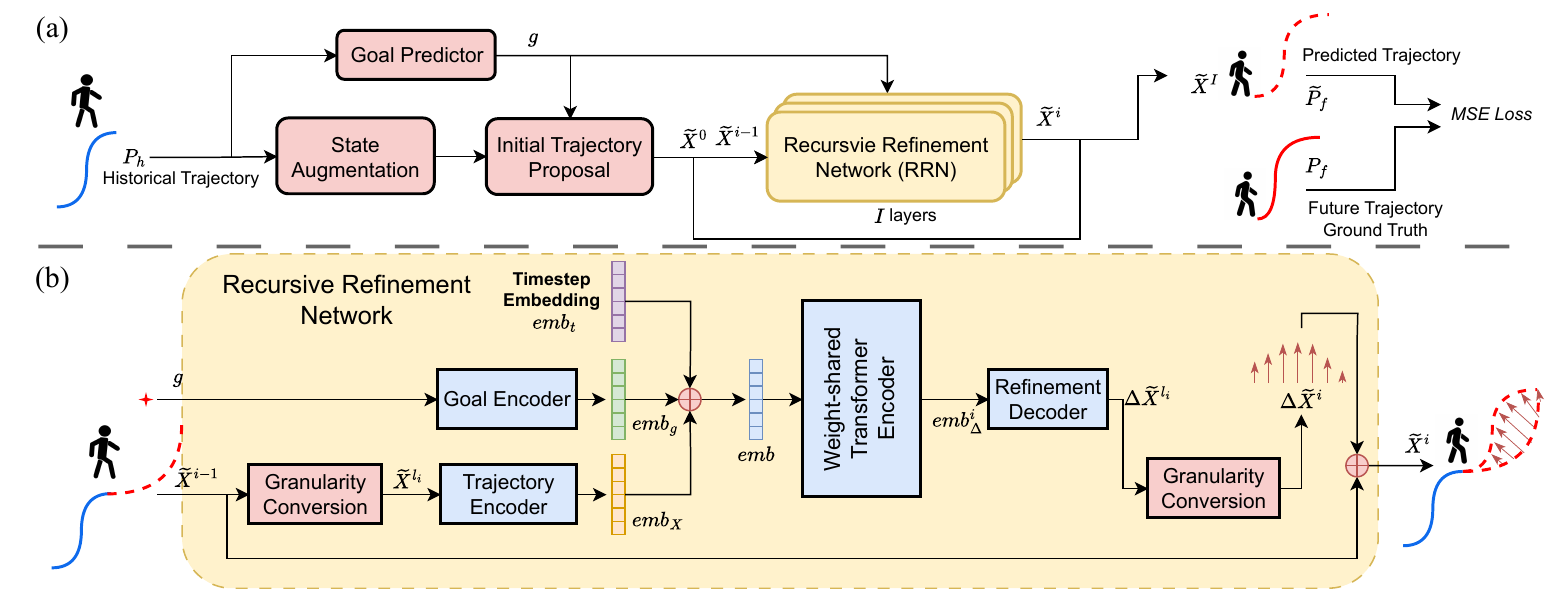}
    \\
    \caption{(a) \textbf{Overall architecture of MGTraj.} MGTraj generates an initial trajectory proposal from the current position and goal prediction. This proposal is recursively refined through multiple recursive refinement networks (RRNs) corresponding to different temporal granularity levels to produce the final future trajectory prediction. The model is trained end-to-end using MSE loss. (b) \textbf{The structure of the RRN.} In each RRN, the input trajectory proposal is first converted to the specific granularity and embedded using a linear layer. This embedded representation is combined with timestep and goal embeddings before being processed by a transformer encoder whose weights are shared across all RRNs. The output of the transformer feeds into separate position and velocity refinement heads. The resulting refinements update the input trajectory proposal, which then acts as the input of the next RRN. Blue blocks denote trainable network components, while red blocks represent non-trainable operations or frozen networks.}
    \label{fig_model}
    \vspace{-14pt}
\end{figure*}

Alternative approaches predict intermediate representations beyond final destinations. For instance, Graph-TERN~\cite{ControlPoint_AAAI_2023} segments the future trajectory using control points spaced at equal intervals. The goal and ultimately the trajectory are predicted by progressively predicting these control points. \textsf{Y}-net~\cite{YNet_2021_ICCV} introduces waypoints prediction in the long-term prediction task, selecting the waypoint as the halfway frame of the prediction horizon. V$^2$-Net~\cite{V2_ECCV_2022} first predicts keypoints and then interpolates the keypoints in the frequency spectrum to reconstruct the complete trajectory.
The promising performance of these models demonstrates the effectiveness of such goal prediction scheme. However, despite these advancements that partition the prediction horizon into different levels of granularity, they remain fundamentally constrained by the two-stage framework with limited granularity exploitation.

\subsection{Multi-Granularity Application}
Multi-granularity (or multi-scale) representations have demonstrated significant value across domains. In object detection, Feature Pyramid Network (FPN)~\cite{FPN_CVPR_2017} integrates high-resolution features with semantic context to improve multi-scale detection. For time-series forecasting, MG-TSD~\cite{MGTSD_ICLR_2024} generates multi-granular data and then leverages them to guide the diffusion model for denoising.
Considering the effectiveness of the multi-granularity idea, pioneer work proposed by Liang \etal~\cite{TPNMS_AAAI_2021} first attempts to apply the temporal pyramid to human trajectory prediction. The proposed TPNMS method extracts the feature and predicts the future trajectory with different granularities by upsampling and downsampling. One disadvantage of this method is that the end position within the segment is used to represent the whole segment during downsampling, which discards too much motion information. MlgtNet~\cite{MlgtNet_TITS_2024} introduces the Gabor transform to convert the history trajectory into the frequency domain and generate different levels of granularity in the spectrum. Similar to V$^2$-Net, the spectral decomposition introduces preprocessing overhead through frequency-domain transformations. More critically, neither approach explicitly incorporates goal-oriented granularity, the coarsest temporal representation fundamental to behavioral intent, which limits their ability to model hierarchical motion semantics.


\section{METHODOLOGY}
\label{sec:proposed method}
\subsection{Overview} 
The human trajectory prediction problem can be formulated as follows:
The entire trajectory is denoted as $X$ with the length $T$. Given the historical state of the targeted agent in the past $T_{h}$ frames $X_h = \{x_{t}| t = 1,2,...,T_{h}\}$, the objective is to obtain $N$ possible future trajectory prediction $X_f$ in the future $T_{f}$ frames, denoted as $\widetilde{X}_f = \{\widetilde{x}_{t}| t = T_h+1, T_h+2,..., T\}$, which should be as close as possible to the ground truth $X_f$.

Fig. \ref{fig_model}(a) presents the overall architecture of MGTraj. The model first utilizes the input historical positions to predict the goal and construct an initial trajectory proposal conditioned on the goal prediction, as elaborated in Section \ref{proposal}. This proposal is then processed by the RRN, which extracts and fuses multi-granular temporal features to iteratively refine the proposal, as depicted in Fig. \ref{fig_model}(b). The RRN's architecture and refinement mechanism are elaborated in Section \ref{model}. Finally, an end-to-end training scheme is applied with an auxiliary velocity prediction task, as introduced in \ref{loss}.

\subsection{Goal Predictor and Trajectory Proposal}
\label{proposal}
\textbf{Goal Predictor} Since this work concentrates on integrating multi-granularity into the goal-guided trajectory prediction framework, we pre-train a transformer with the causal mask as the goal predictor, following previous work~\cite{PPT_ECCV_2024}. The goal predictor is pre-trained in two stages: first, it learn to predict the position of the immediate next frame based on historical trajectory embedding. Subsequently, in the second stage, a learnable goal embedding is concatenated with historical trajectory embedding and serves as the input to the transformer. The final output token of the transformer is then decoded to generate a multi-modal distribution of potential goal predictions, denoted as $G = \{g^n=p_T^n|n=1,2,..., N\}$ for multi-modal prediction. For simplicity, the notation of modality will be ignored in the rest of the paper. 

\textbf{Initial Trajectory Proposal} 
Motivated by previous work~\cite{Trajectron++_2020_ECCV, GDTS_IROS_2025}, the position $P \in \mathbb{R}^{T\times2}$ is augmented with velocity $V \in \mathbb{R}^{T\times2}$ to obtain the augmented state $X \in \mathbb{R}^{T\times4}$.
For each modality, the initial trajectory proposal $\widetilde{X}^0$ can be obtained by linear interpolation based on goal prediction $p_T$, as shown in the following equations:
\begin{equation}
\label{pos_interpolation}
p_t = \begin{cases}
             \ \bar{p}_t, & 1\leq t \leq T_{obs} \\  
             \ \bar{p}_{T_{obs}} + \frac{t-T_{obs}}{T-T_{obs}}(p_T-\bar{p}_{T_{obs}}),& T_{obs} < t \leq T_{}
\end{cases}
\end{equation} 

\begin{equation}
\label{vel_interpolation}
v_t = \begin{cases}
            \ \bar{v}_t, & 1\leq t \leq T_{obs} \\
             \ \frac{p_T-\bar{p}_{T_{obs}}}{T-T_{obs}},& T_{obs} < t \leq T_{}
\end{cases}
\end{equation}

\subsection{Transformer-based Recursive Refinement Network}
\label{model}
\textbf{Granularity Conversion}  We denote the number of granularity levels as $I$ and the granularity level list as $GL = \{l_i|i=1,..., I\}$ from coarse to fine. Different from V$^2$-Net~\cite{V2_ECCV_2022} and MlgtNet~\cite{MlgtNet_TITS_2024}, where the number of spectral keypoints or segment sizes must be exactly divisible by the future horizon $T_f$, our method requires that the granularity level $l$ be divisible by the full observed trajectory length $T$. This relaxation expands the range of possible values for $l_i$ under our experiment setting, offering larger flexibility in designing multi-granularity representations. The trajectory proposal $\widetilde{X}$ needs to be first converted into the $i$-th granularities so that the $i$-th refinement network could capture the feature of this granularity. Specifically, the temporal dimension of the trajectory proposal $X$ will be divided into $l_i$ segments equally, and the $i$-th granularity of $\widetilde{X}$ is $X^{l_i} \in \mathbb{R}^{\frac{T}{l_i}\times4l_i}$. 

\textbf{Refinement Prediction} After receiving the $i$-th granularity of trajectory proposal $X^{l_i}$, a transformer-based recursive refinement network (RRN) is applied to capture the feature and predict the refinement. As shown in Fig. \ref{fig_model}, a trajectory encoder is first leveraged to extract the embedding of the trajectory proposal $emb_{X}$, and the goal embedding $emb_g$ is obtained using a goal encoder. Similar to~\cite{Transformer_Neurips_2017}, we also incorporate the chronological order into embedding by applying sinusoidal position encoding to the timestep range from 1 to $T$. The temporal position embedding is denoted as $emb_t$. The final input of the transformer will be the addition of these three embeddings:
$$emb = emb_g + emb_t + emb_{X}$$
Receiving the input embedding $emb$, the transformer encoder performs self-attention along the temporal dimension to capture the temporal dynamics of the pedestrian trajectory with the $i$-th granularity. Crucially, we implement weight-shared scheme of the transformers across different RRNs. Besides implicitly providing the model with the prior knowledge that these inputs are the same trajectory with different granularities, this design also maintains parameter efficiency by ensuring model complexity remains bounded despite the increasing number of granularity levels.

After obtaining the features from the transformer, a refinement decoder is employed to predict the refinement of the state $\Delta \widetilde{X}^{l_i}$. This refinement $\Delta \widetilde{X}^{l_i}$ is subsequently converted back to the finest granularity as $\Delta \widetilde{X}^i \in \mathbb{R}^{T\times 4}$. The trajectory proposal is then updated as $\widetilde{X}^{i} = \widetilde{X}^{i-1} + \Delta \widetilde{X}^{i}$, preparing it for further refinement in the next RRN.

The granularity conversion and refinement prediction will be repeated recursively until the refinement of the finest granularity is obtained, and the predicted final state is: $$\widetilde{X}=\widetilde{X}^0+\Delta \widetilde{X}^1 + \cdots + \Delta \widetilde{X}^I$$
Since the ultimate objective is to predict only the position of the future $T_f$ frames, we separate the final position prediction $\widetilde{P}_f$ out from state $\widetilde{X}$.

\begin{table}[tbp]
\caption{\textbf{Quantitative results on Stanford Drone Dataset (SDD).} Bold number indicates the \textbf{best} goal-guided model.} 
\centering
\resizebox{0.9\columnwidth}{!}{
\begin{tabular}{l|c|c|c}
    \toprule
    \textbf{Method} & Map Information & \textbf{ADE$_{20}$} & \textbf{FDE$_{20}$}\\
    \cmidrule{1-4}
    V$^2$-Net (2022)~\cite{V2_ECCV_2022} & $\checkmark$ & 7.12 & 11.39 \\
    MlgtNet (2024)~\cite{MlgtNet_TITS_2024} & $\checkmark$& 6.91 & 11.04 \\
    \cmidrule{1-4}
    PECNet (2020)~\cite{PECNet_2020_ECCV} & & 9.96 & 15.88\\
    \textsf{Y}-net (2021)~\cite{YNet_2021_ICCV} & $\checkmark$ & 7.85 & 11.85\\
    Goal-SAR (2022)~\cite{GoalSAR_2022_CVPRW} & $\checkmark$ & 7.75 & 11.83 \\
    Graph-TERN (2023)~\cite{ControlPoint_AAAI_2023} & & 8.43 & 14.26 \\
    GDTS (2025)~\cite{GDTS_IROS_2025} & $\checkmark$ & 7.42 & 11.57 \\
    PPT (2024)~\cite{PPT_ECCV_2024} & & 7.03 & 10.65 \\
    \cmidrule{1-4}
    MGTraj (Ours) & & \textbf{6.98} & \textbf{10.55} \\
    \bottomrule
\end{tabular}
}
\label{tab:SDD}
\end{table}

\subsection{Training}
\label{loss}
Our model is trained in an end-to-end manner. Considering the objective, we want to minimize the mean square error (MSE) between the prediction and the ground truth of the position.
\begin{equation}  
\mathcal{L}_p = MSE(\widetilde{p},p)
\end{equation} 
To encourage the model to extract motion dynamics, we also perform the MSE loss between the prediction and the ground truth of the velocity.
\begin{equation}  
\mathcal{L}_v = MSE(\widetilde{v},v)
\end{equation} 
And the total loss is computed as:
\begin{equation}  
\mathcal{L} = \mathcal{L}_p + \lambda_v \mathcal{L}_v
\end{equation} 
where $\lambda_v$ is the loss weight of the velocity.

\section{RESULTS}

\begin{table*}[thbp]
\caption{\textbf{Quantitative results on ETH/UCY dataset.} Bold and underlined number indicates the \textbf{best} and the \underline{second-best} goal-guided model.} 
\centering
\resizebox{1.9\columnwidth}{!}{
\begin{tabular}{l|c|c c |c c |c c |c c |c c |c c}
    \toprule
    ~&\multirow{2}{*}[-1ex]{\shortstack{Map \\Information}}&\multicolumn{2}{c|}{\textbf{ETH}} & \multicolumn{2}{c|}{\textbf{HOTEL}} &  \multicolumn{2}{c|}{\textbf{UNIV}} &  \multicolumn{2}{c|}{\textbf{ZARA1}} & \multicolumn{2}{c|}{\textbf{ZARA2}} & \multicolumn{2}{c}{\textbf{AVG}} \\
    \cmidrule{3-14}
    & &\multicolumn{1}{c}{\textbf{ADE$_{20}$}} & \multicolumn{1}{c|}{\textbf{FDE$_{20}$}} &
    \multicolumn{1}{c}{\textbf{ADE$_{20}$}} &  \multicolumn{1}{c|}{\textbf{FDE$_{20}$}} &
    \multicolumn{1}{c}{\textbf{ADE$_{20}$}} &  \multicolumn{1}{c|}{\textbf{FDE$_{20}$}} &
    \multicolumn{1}{c}{\textbf{ADE$_{20}$}} &  \multicolumn{1}{c|}{\textbf{FDE$_{20}$}} &
    \multicolumn{1}{c}{\textbf{ADE$_{20}$}} &  \multicolumn{1}{c|}{\textbf{FDE$_{20}$}} &
    \multicolumn{1}{c}{\textbf{ADE$_{20}$}} &  \multicolumn{1}{c}{\textbf{FDE$_{20}$}}
    
    \\
    \cmidrule{1-14}
    TPNMS (2021)~\cite{TPNMS_AAAI_2021} & & 0.52 & 0.89 & 0.22 & 0.39 & 0.55 & 1.13 & 0.35 & 0.70 & 0.27 & 0.56 & 0.38 & 0.73\\
    V$^2$-Net (2022)~\cite{V2_ECCV_2022} & $\checkmark$ & 0.23 & 0.37 & 0.11 & 0.16 & 0.21 & 0.35 & 0.19 & 0.30 & 0.14 & 0.24 & 0.18 & 0.28 \\
    MlgtNet (2024)~\cite{MlgtNet_TITS_2024} & $\checkmark$ & 0.24 & 0.36 & 0.10 & 0.14 & 0.17 & 0.29 & 0.17 & 0.28 & 0.13 & 0.22 & 0.16 & 0.26 \\
    \cmidrule{1-14}
    PECNet (2020)~\cite{PECNet_2020_ECCV} & & 0.54&0.87 & 0.18&0.24 & 0.35&0.60 & 0.22&0.39 & 0.17&0.30 & 0.29&0.48\\
    \textsf{Y}-net (2021)~\cite{YNet_2021_ICCV} & $\checkmark$ & \textbf{0.28}&\textbf{0.33} & \textbf{0.10}&\textbf{0.14} & 0.24&0.41 & \underline{0.17}&\underline{0.27} & \underline{0.13}&0.22 & \textbf{0.18}&\textbf{0.27} \\
    Goal-SAR (2022)~\cite{GoalSAR_2022_CVPRW} & $\checkmark$ & \textbf{0.28}&\underline{0.39} & 0.12&0.17 & 0.25&0.43 & \underline{0.17}&\textbf{0.26} &  0.15&0.22 &  \underline{0.19}&\underline{0.29}\\
    Graph-TERN (2023)~\cite{ControlPoint_AAAI_2023} & & 0.42 & 0.58 & 0.14 & 0.23 & 0.26 & 0.45 & 0.21 & 0.37 & 0.17 & 0.29 & 0.24 & 0.38 \\
    GDTS (2025)~\cite{GDTS_IROS_2025} & $\checkmark$& 0.31 & 0.48 & 0.13 & 0.18 & 0.27 & 0.49 & 0.19 & 0.29 & 0.15 & 0.24 & 0.21 & 0.33 \\
    PPT (2024)~\cite{PPT_ECCV_2024} & & 0.36&0.51 & \underline{0.11}&\underline{0.15} & \textbf{0.22}&\underline{0.40} & \underline{0.17}&0.30 & \textbf{0.12}&\underline{0.21} & 0.20&0.31 \\
    \cmidrule{1-14}
    MGTraj (Ours) & & \textbf{0.28}&0.45 & 0.13&0.16 & \textbf{0.22}&\textbf{0.37} & \textbf{0.16}&0.29 & \underline{0.13}&\textbf{0.20} & \textbf{0.18}&\underline{0.29} \\
    \bottomrule
\end{tabular}
}
\label{tab:eth_ucy}
\end{table*}

\subsection{Experiments and Datasets}
\label{sec:experiments}
\textbf{Datasets}
We conduct experiments on the ETH/UCY dataset~\cite{ETH_2009_ICCV, UCY_2007} and the Stanford Drone Dataset (SDD)~\cite{SDD_2016_ECCV} to evaluate the performance of our model. The ETH/UCY dataset consists of five scenes: ETH, Hotel, Univ, Zara1, and Zara2, with a total of 1,536 pedestrians. Following the common leave-one-scene-out protocol~\cite{SocialGAN_2018_CVPR, YNet_2021_ICCV}, models are trained on four scenes and evaluated on the remaining held-out scene. a large-scale dataset captured across a university campus, containing 5,232 labeled pedestrians in eight different scenes. We use the same dataset split as several recent works \cite{GoalSAR_2022_CVPRW, PECNet_2020_ECCV}. All datasets are downsampled to 2.5 FPS. The observation length is $T_h = 8$ (3.2s) while the prediction length is $T_f = 12$ (4.8s).

\textbf{Evaluation Metrics}
We leverage Average Displacement Error (ADE) and Final Displacement Error (FDE) as the evaluation metrics. ADE calculates the average Euclidean distance between the ground truth and the predicted positions across all time steps from $T_h+1$ to $T$, while FDE only measures Euclidean distance at the last timestep $T$. Following the common setting in prior works~\cite{YNet_2021_ICCV, GoalSAR_2022_CVPRW, GDTS_IROS_2025, PPT_ECCV_2024}, we generate $N=20$ predictions of future trajectory, and report the best-of-$N$ ADE and FDE of the predictions, denoted as ADE$_N$ and FDE$_N$, which measure the minimum ADE and FDE among all predicted possible future trajectories, respectively.

\textbf{Implementation Details}
We train our model using ADAM optimizer with a batch size of $512$ for $500$ epochs on an NVIDIA RTX 3080Ti GPU with PyTorch implementation, and the goal predictor adopted from PPT~\cite{PPT_ECCV_2024} is pretrained. The initial learning rate is $1\times10^{-3}$, and cosine annealing scheme is applied. The trajectory encoder, the goal encoder, and the refinement decoder are all linear layers with GELU activation function. The embedding size of the one-layer transformer encoder is set to be $64$ and the attention head is $8$. The weight of the velocity MSE loss $\lambda_v=5$. We select the number of granularity levels $I$ to be 4, and the granularity level list is $GL = \{l_1=10,l_2=4,l_3=2,l_4=1\}$.

\subsection{Quantitative Results}
\textbf{Baseline}
We compare our model to different goal-guided baselines, including PECNet~\cite{PECNet_2020_ECCV}, \textsf{Y}-net~\cite{YNet_2021_ICCV}, Goal-SAR~\cite{GoalSAR_2022_CVPRW}, Graph-TERN~\cite{ControlPoint_AAAI_2023}, GDTS~\cite{GDTS_IROS_2025}, and PPT~\cite{PPT_ECCV_2024}. Despite open-source, NSP-SFM~\cite{NSPSFM_2022_ECCV} is excluded due to its scene-specific hyperparameter tuning on the ETH/UCY dataset, which introduces unfair advantages. We also include results from trajectory prediction methods employing multi-granularity representations, specifically TPNMS~\cite{TPNMS_AAAI_2021}, V$^2$-Net~\cite{V2_ECCV_2022}, and MlgtNet~\cite{MlgtNet_TITS_2024}. Inputs vary across these models: while some methods utilize RGB maps as additional input, others predict future trajectories solely based on historical trajectory data.

\textbf{Discussion} Table \ref{tab:SDD} shows the experimental results on the SDD, with the best metrics of the goal-guided methods highlighted in bold. MGTraj achieves the best performance among all goal-guided baselines in both ADE$_{20}$ and FDE$_{20}$ metrics. When compared to multi-granularity approaches, our method achieves a small ADE$_{20}$, which is only higher than MLgtNet, despite using only historical trajectory input. By explicitly leveraging the goal prediction as the coarsest representation, our method surpasses all of the other previous multi-granularity methods in FDE$_{20}$ metric, including MlgtNet.

Furthermore, we also report the results of experiments on the ETH/UCY dataset in Table \ref{tab:eth_ucy}. MGTraj achieves competitive accuracy with the state-of-the-art human trajectory prediction methods, including MlgtNet and \textsf{Y}-net, even without leveraging map information. Notably, \textsf{Y}-net leverage test-time-sampling-trick in inference, which is computationally costly. With the adoption of the goal predictor from PPT~\cite{PPT_ECCV_2024}, our model outperforms it across nearly all scenes, which indicates the benefit of integrating multi-granularity into goal-guided methods.

The consistently superior performance in both datasets demonstrates the effectiveness of our proposed approach. By employing multiple RRNs corresponding to different temporal granularity levels, our model recursively extracts and integrates trajectory features from coarse to fine granularity. Additionally, our lightweight model contains approximately 0.6 million trainable parameters, which is roughly one-fourth of PPT's parameter count. 

\begin{table}[t]
\caption{\textbf{Abaltion study of Different $GL$ of MGTraj.} Bold number indicates the \textbf{best}.} 
\centering
\resizebox{0.6\columnwidth}{!}{
\begin{tabular}{c|c|c}
    \toprule
    \textbf{$GL$} & \textbf{ADE$_{20}$} & \textbf{FDE$_{20}$}\\
    \cmidrule{1-3}
    \{1\} & 7.12 & 10.72 \\
    \{10\} & 7.06 & 10.68 \\
    \cmidrule{1-3}
    \{2,1\} & 7.10 & 10.72\\
    \{4,2,1\} & 7.08 & 10.70\\
    \{1,1,1,1\} & 7.13 & 10.67\\
    \{10,10,10,10\} & 7.14 & 10.66 \\
    \cmidrule{1-3}
    \{10,4,2,1\} (Ours) & \textbf{6.98} & \textbf{10.55} \\
    \bottomrule
\end{tabular}
}
\label{tab:ablation_1}
\end{table}

\begin{figure}[t]
    \scriptsize
    \setlength{\tabcolsep}{1.5pt}
    \centering
    \includegraphics[width=0.9\linewidth]{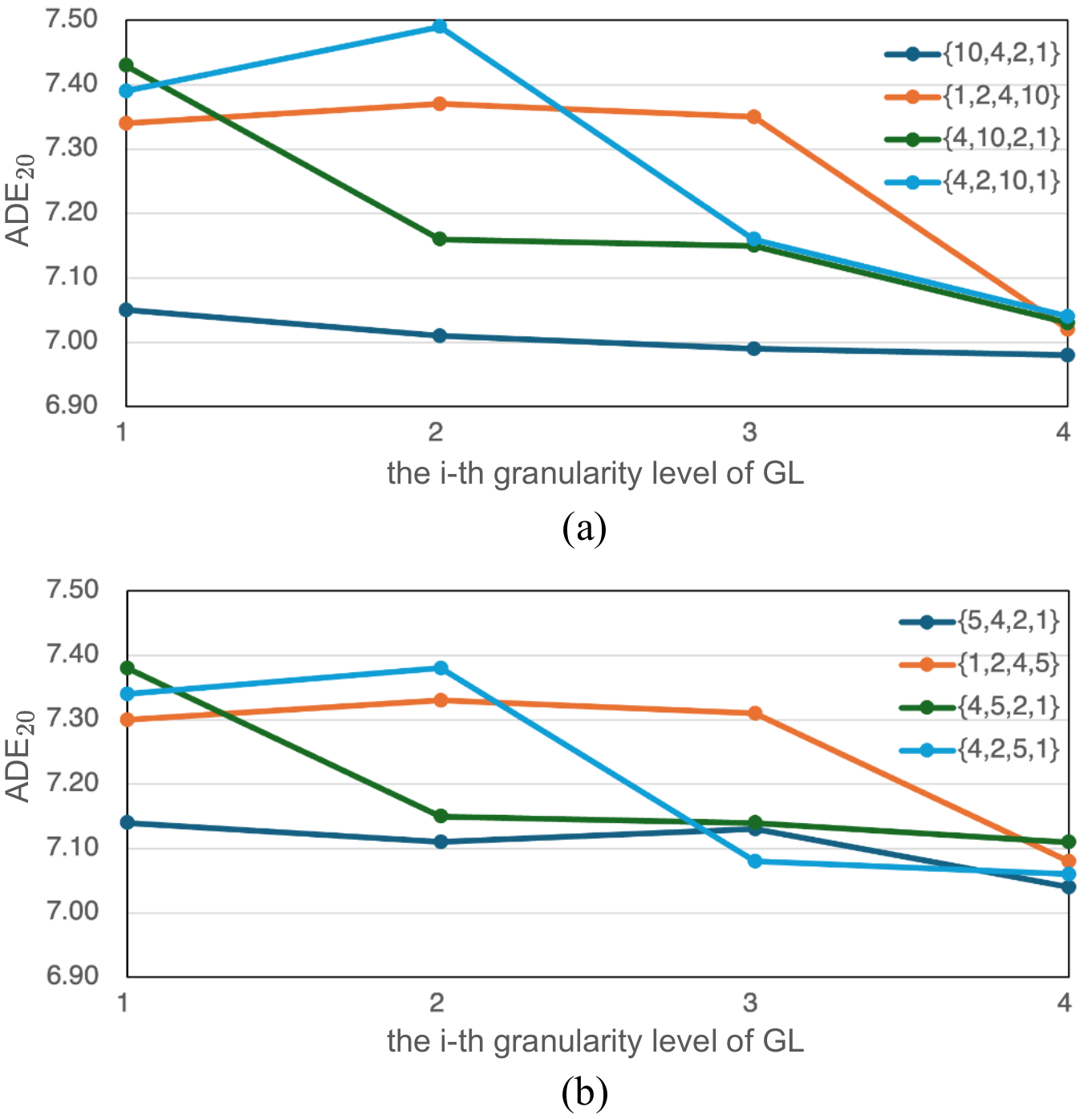}
    \\
    \caption{Performance of models with different $GL$ on SDD, measured with ADE$_{20}$. The RRN with granularity $l=10$ contributes the most in improving accuracy when $GL$ contains $l=10$, and the RRN with granularity $l=5$ contributes the most in improving accuracy when $GL$ contains $l=5$.}
    \label{fig_gl}
    \vspace{-14pt}
\end{figure}

\subsection{Ablation Study}
\textbf{Granularity Level List}
SDD is selected for our ablation studies due to its larger scale and more representative motion. We first conduct experiments on different granularity level sets $GL$ to verify the effectiveness of our multi-granularity design, as summarized in Table \ref{tab:ablation_1}. When degrading our multi-granularity model into the single-granularity setting, for example, $GL=\{1\}$ or $\{10\}$, the performance also degraded. We then incrementally add coarser granularity levels to $GL=\{1\}$. The results of $GL=\{2, 1\}$, $\{4, 2, 1\}$, and $\{10, 4, 2, 1\}$ demonstrate that the prediction accuracy improves with more granularity levels considered in trajectory prediction. Furthermore, stacking the finest refinement network four times with $GL=\{1,1,1,1\}$ yields comparable ADE$_{20}$ to $GL=\{1\}$, while $GL=\{10,10,10,10\}$ even performs worse. This indicates that rather than simply increasing network depth or parameters, the integration of features from multiple granularity levels is the key factor in improving prediction accuracy. 

Notably, when adding granularity level $l=10$ to $GL=\{4, 2, 1\}$ (yielding $GL =\{10, 4, 2, 1\}$), the model achieves relatively larger improvement in both ADE$_{20}$ and FDE$_{20}$. This observation raises two questions: \textit{a. Does the granularity level $l=10$ contribute the most to the refinement of the proposal? b. If so, what is the possible underlying reason?} Different from TPNMS~\cite{TPNMS_AAAI_2021}, our training uses only the MSE loss of the final output without multiple supervision of different granularity levels, allowing the model to automatically distribute the importance of each granularity.

To answer the first question, we experiment with different orderings of the same granularity levels: $GL=\{1, 2, 4, 10\}$, $\{4, 10, 2, 1\}$, and $\{4, 2, 10, 1\}$. We compute the ADE$_{20}$ after each RRN to evaluate the contribution of each granularity. As shown in Fig. \ref{fig_gl}(a), the RRN with $l=10$ leads to the largest improvement in accuracy, compared with RRNs with other granularity levels. A potential explanation is that $l=10$ divides the whole trajectory of length $T=20$ into two segments, enabling higher-level motion understanding. Meanwhile, the first segment contains both history observation and future prediction, implicitly encouraging motion consistency between history and prediction, thereby improving the accuracy. This is supported by the superior performance of $GL = \{10\}$ over $GL = \{1\}$, as shown in Table~\ref{tab:ablation_1}. To further validate this hypothesis, we replace $l=10$ with $l=5$ in $GL$, which also includes a segment spanning both history and future timesteps. As illustrated in Fig. \ref{fig_gl}(b), $l=5$ contributes most significantly to performance improvement, consistent with the observation when $l=10$. However, since $l=5$ is close to $l=4$, it may hinder the model's ability to understand motion and capture features in a sufficiently coarse resolution. Thus, models using $l=5$ achieve suboptimal ADE$_{20}$ compared to our model using $l=10$.

In addition, the result in Fig. \ref{fig_gl} demonstrates that the order of granularity level also has a slight yet noticeable influence on the model performance. Gradually refining the trajectory proposal from the coarsest to the finest granularity proves to be more natural and achieves lower ADE$_{20}$ and FDE$_{20}$ compared to other ordering strategies.


\begin{figure}[t]
    \scriptsize
    \setlength{\tabcolsep}{1.5pt}
    \centering
    \includegraphics[width=0.8\linewidth]{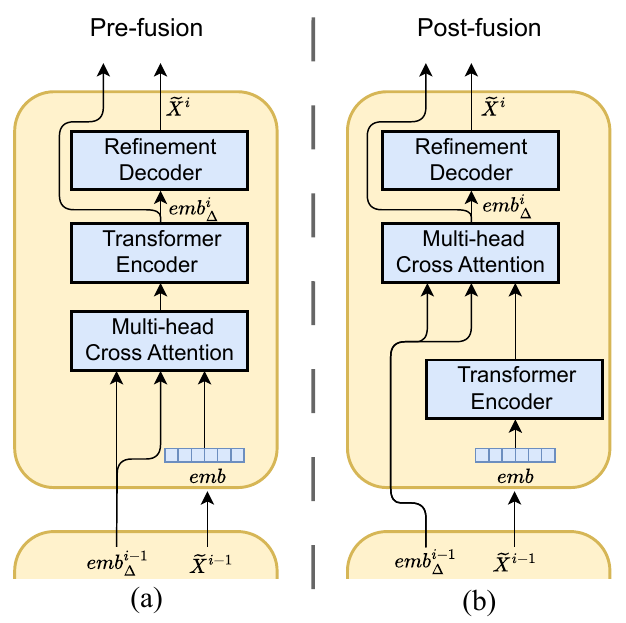}
    \\
    \caption{Illustration of the $i$-th RRN using two alternatives for feature fusion: (a) pre-fusion and (b) post-fusion, in which transformer encoder weights are not shared across RRNs. Certain operations and components of RRN are omitted for clarity.}
    \label{fig_fusion}
    \vspace{-14pt}
\end{figure}

\begin{figure*}[t]
    \scriptsize
    \setlength{\tabcolsep}{1.5pt}
    \centering
    \includegraphics[width=0.9\linewidth]{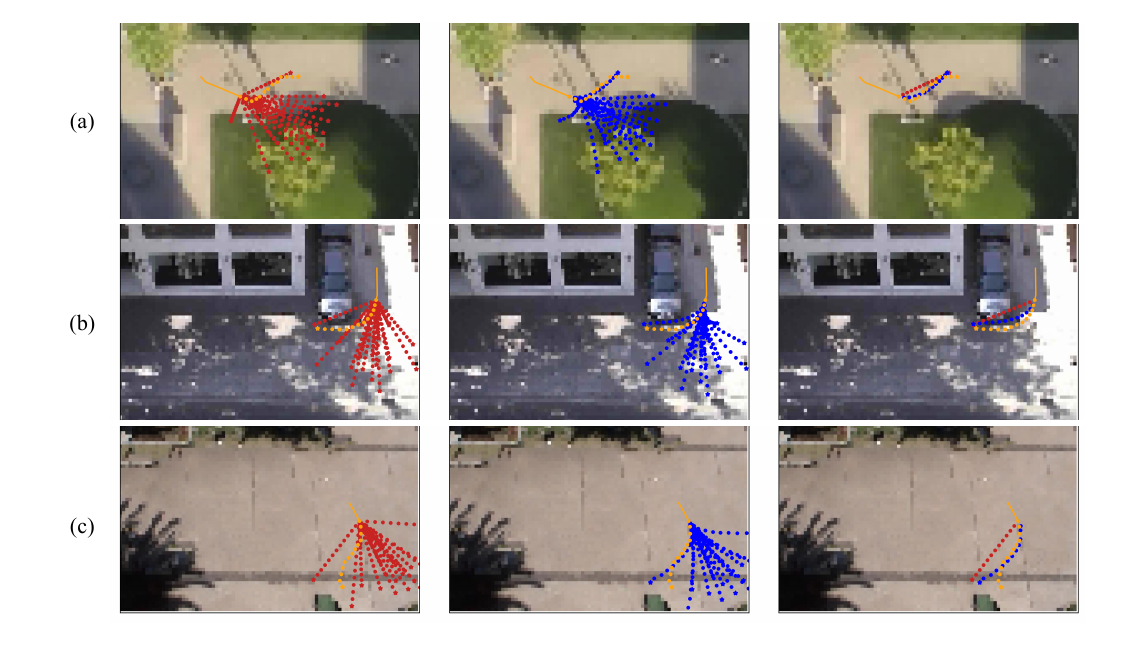}
    \\
    \caption{Visualization comparison of initial trajectory proposal and final prediction of MGTraj in three different scenarios. The line indicates the historical trajectory. The ground truth future trajectory is in yellow, the goal predictions and the corresponding initial trajectory proposals are in red, and the final predictions are in blue. The goals are highlighted as star points. In the right column, the initial proposal and the final trajectory prediction closest to the ground truth are selected for better comparison.}
    \label{fig_qualitative}
    \vspace{-14pt}
\end{figure*}

\textbf{Feature Fusion}
To verify that our weight-shared feature fusion strategy enhances the ability of the model to effectively capture and fuse the features across different temporal granularity levels, we first evaluate the importance of feature fusion by employing separate transformer encoders for each RRN without sharing weights. In this case, there will be no explicit feature fusion in the model. Furthermore, we propose two model variants to explore alternative fusion approaches using cross-attention: the embedding from the current granularity serves as the query, while the embedding from the previous level serves as the key and value. As shown in Fig. \ref{fig_fusion}, in the pre-fusion variant, the embeddings are first fused and then processed through the transformer. And post-fusion variant performs the cross-attention feature fusion after the embeddings have been processed by their respective transformer-based encoders. Results are presented in Table \ref{tab:ablation_2}. The degraded performance of the model without feature fusion demonstrates the importance of feature fusion. The pre-fusion and post-fusion variants achieve marginal improvement in accuracy, while our weight-shared fusion strategy outperforms both alternatives, which indicates the superiority of our feature fusion strategy in multi-granularity methods. By implicitly providing the prior knowledge that inputs with different granularity levels represent the same underlying trajectory, our weight-shared strategy optimizes feature fusion effectiveness. Concurrently, this strategy benefits the parameter efficiency of the model without compromising performance.

\begin{table}[t]
\caption{\textbf{Ablation study of MGTraj on features fusion.} Bold number indicates the \textbf{best}.} 
\centering
\resizebox{0.7\columnwidth}{!}{
\begin{tabular}{c|c|c}
    \toprule
    \textbf{Feature Fusion} & \textbf{ADE$_{20}$} & \textbf{FDE$_{20}$}\\
    \cmidrule{1-3}
    no fusion &  7.12 & 10.68 \\
    pre-fusion &  7.08 & 10.66 \\
    post-fuison &  7.05 & 10.63 \\
    \cmidrule{1-3}
    weight-shared (Ours) &  \textbf{6.98} & \textbf{10.55} \\
    \bottomrule
\end{tabular}
}
\label{tab:ablation_2}
\end{table}

\begin{table}[t]
\caption{\textbf{Ablation study of MGTraj on velocity augmentation and loss weight $\lambda_v$.} Bold number indicates the \textbf{best}.} 
\centering
\resizebox{0.65\columnwidth}{!}{
\begin{tabular}{c|c|c|c}
    \toprule
    \textbf{Vel. Aug.} & \textbf{$\lambda_v$} & \textbf{ADE$_{20}$} & \textbf{FDE$_{20}$}\\
    \cmidrule{1-4}
     & - & 7.09 & 10.71 \\
    $\checkmark$ & 0 & 7.06 & 10.67 \\
    $\checkmark$ & 1 & 7.01 & 10.59 \\
    $\checkmark$ & 10 & 7.04 & 10.61 \\
    \cmidrule{1-4}
    $\checkmark$ & 5 &\textbf{6.98} & \textbf{10.55} \\
    \bottomrule
\end{tabular}
}
\label{tab:ablation_3}
\end{table}

\textbf{Auxiliary Task}
We also conduct an ablation study to validate that velocity augmentation and velocity prediction acting as auxiliary task actually improve the prediction accuracy. As indicated in Table \ref{tab:ablation_3}, removing the velocity augmentation leads to a noticeable degradation in the model performance, confirming its importance in enhancing prediction accuracy. Furthermore, we investigate the influence of the velocity loss weight $\lambda_v$ on prediction performance. Notably, when $\lambda_v=0$, the model is trained solely by position loss $\mathcal{L}_p$. Although velocity augmentation alone yields some improvement, the absence of explicit velocity prediction hinders the model from achieving better performance. When $\lambda_v \neq 0$, the model's accuracy first increases and then decreases as $\lambda_v$ becomes large. Based on these observations, we choose $\lambda_v = 5$ as the weight for the velocity MSE loss. By explicitly predicting and minimizing the MSE of the velocity to incorporate soft constraints through joint optimization, the model learn the motion consistency between position and velocity. Fewer abnormal out-of-distribution trajectories will be predicted, such as those with impossible high velocity, thereby the accuracy of prediction will be improved.

\subsection{Qualitative Results}
Fig. \ref{fig_qualitative} illustrates MGTraj's effectiveness through trajectory visualizations across diverse scenarios.
The left column displays goal predictions and corresponding initial trajectory proposals. The middle column presents final predictions after multi-granular refinement through Recursive Refinement Networks (RRNs). For better comparison, the right column presents the initial proposal and the final prediction closest to the ground truth. The qualitative result reveals that MGTraj refines initial proposals into smoother trajectories that better reflect natural human locomotion patterns, as shown in scenario (a). In scenario (b), where a pedestrian navigates around a parked vehicle, the refined trajectory demonstrates adaptive obstacle avoidance behavior, while the initial proposal intersects the obstacle region. Notably in scenario (c), despite deviation in both goal prediction and initial proposal from ground truth, our refinement process substantially improves trajectory accuracy: the first eight future frames align closely with the ground truth, while the final goal prediction is also refined toward the ground truth goal.

\section{CONCLUSIONS}
In this paper, we introduce MGTraj, a novel goal-guided multi-granularity model for human trajectory prediction. With multiple RRNs corresponding to different levels of granularity, MGTraj refines the initial linearly-interpolated trajectory proposal and successfully bridges the gap between the coarsest goal prediction and the finest trajectory completion, with the enhancement of feature fusion and auxiliary velocity prediction task.
Extensive experiments on real-world datasets, including ETH/UCY and Stanford Drone Dataset, demonstrate the effectiveness of MGTraj, with state-of-the-art performance among goal-guided trajectory prediction methods across multiple real-world datasets. 

\textbf{Future work} Our model could function as a refinement module for existing human trajectory prediction methods, which will be investigated in the future. Furthermore, we plan to extend this multi-granularity framework to vehicle trajectory prediction in the context of autonomous driving.








\renewcommand*{\bibfont}{\footnotesize}
\printbibliography

\end{document}